\documentclass[english,ams,yap]{iitpinfo}
\usepackage[utf8]{inputenc}
\usepackage{hyperref}
\usepackage{amsmath}
\usepackage{array}
\usepackage{booktabs}
\usepackage{amssymb,xcolor,stackengine,graphicx}

\newcommand{\vecX}{\mathbf{x}}

\newcommand{\vecZ}{\mathbf{z}}

\newcommand{\vecM}{\boldsymbol{\mu}}

\VolumeNo{24}
\IssueNo{1}
\YearOfIssue{2024}
\CopyrightYear{2024}
\CopyrightedAuthors{Erlygin}

\makeatletter
\renewcommand{\inst}[1]{%
  \unskip
  $^{\@instsymbollist{#1}}$%
}
\newcommand{\@instsymbollist}[1]{%
  \@for\next:=#1\do{%
    \@instsymbol{\next}%
    \ifx\next\@lastofloop\else\,\fi
  }%
}
\makeatother

\title{Uncertainty Estimation for the Open-Set Text Classification systems\thanks{The research was supported by the Russian Science Foundation grant No. 25-11-00355}}

\author{L.\,A.\,Erlygin\inst{1}, A.\,A.\,Zaytsev\inst{1} \inst{2}}

\institute{Skolkovo Institute of Science and Technology (Skoltech), Moscow, Russia \and Risk Management, Sber, Moscow, Russia
}

\received{Received September 15, 2023}

\titlerunning{Uncertainty Estimation for the Open-Set Text Classification systems}

\authorrunning{ERLYGIN}

\Rubric{MACHINE LEARNING AND PATTERN RECOGNITION}

\begin{document}

\maketitle
\begin{abstract}
Accurate uncertainty estimation is essential for building robust and trustworthy recognition systems.
In this paper, we consider the open-set text classification (OSTC) task --- and uncertainty estimation for it.
For OSTC a text sample should be classified as one of the existing classes or rejected as unknown.
% Several practical applications fall under this problem statement: authorship identification, intent classification, and topic recognition.
% Despite the relevance of these problems, uncertainty estimation in OSTC receives little attention in the literature.
To account for the different uncertainty types encountered in OSTC, we adapt the Holistic Uncertainty Estimation (HolUE) method for the text domain.
Our approach addresses two major causes of prediction errors in text recognition systems: text uncertainty that stems from ill formulated queries and gallery uncertainty that is related the ambiguity of data distribution.

% TODO gallery uncertainty - как-то пояснить
By capturing these sources, it becomes possible to predict when the system will make a recognition error.
We propose a new OSTC benchmark and conduct extensive experiments on a wide range of data, utilizing the authorship attribution, intent and topic classification datasets.
HolUE achieves 40-365\% improvement in Prediction Rejection Ratio (PRR) over the quality-based SCF baseline across datasets: 365\% on Yahoo Answers (0.79 vs 0.17 at FPIR 0.1), 347\% on DBPedia (0.85 vs 0.19), 240\% on PAN authorship attribution (0.51 vs 0.15 at FPIR 0.5), and 40\% on CLINC150 intent classification (0.73 vs~0.52).
We make public our code and protocols \url{https://github.com/Leonid-Erlygin/text_uncertainty.git} 
% TODO add links to (github) code in abstract
\end{abstract}


\section{INTRODUCTION}
% problem setting
The Open-Set Recognition (OSR) problem arises in many practical applications spanning image, audio, and text domains. 
In this setting, a recognition system maintains a gallery of known classes, representing the set of identities or categories enrolled during training.
For every incoming data sample, referred to as a probe, the system must decide whether the sample belongs to one of the known classes in the gallery and provide its class identifier, or reject the probe as unknown~\cite{scheirer2013toward, face_handbook}.
This formulation is critical for deploying robust systems in open-world scenarios where encountering unseen categories is inevitable.
Several Natural Language Processing (NLP) tasks can be naturally formulated as text-based OSR problems, including intent classification, authorship attribution, and topic recognition~\cite{osr_in_text_review}.
For example, in intent classification, a conversational agent must handle user queries unknown to the system without forcing an incorrect classification, while in authorship attribution, the system must verify whether a document originates from a known writer or an impostor.
Similarly, topic classification systems often encounter articles that do not fit predefined categories, requiring robust rejection mechanisms to maintain system integrity.

% research gap
Despite the prevalence of these tasks, existing research in the text domain focuses predominantly on improving recognition accuracy or enhancing out-of-distribution (OOD) detection capabilities~\cite{osr_in_text_review}.
Methods typically optimize embedding discriminability to separate known from unknown classes, aiming to minimize error rates such as False Acceptance or False Rejection. However, high accuracy alone does not guarantee system trustworthiness in risk-sensitive applications. 
A robust recognition system must be capable of estimating the uncertainty of its predictions to determine when to abstain from making a decision.
When the uncertainty associated with a particular probe is high, the system can defer the decision to a human operator or request additional data, thereby preventing potential errors~\cite{ABDAR2021243}.
While uncertainty estimation has gained traction in computer vision, it receives considerably less attention in text-based OSR systems.
To our knowledge, no prior work has estimated uncertainty in text classification systems specifically for error detection, rather than solely for improving classification accuracy.

% our method. Here I deliberate on the importance of capturing all sources of uncertainty

\begin{figure*}
    \includegraphics[width=1\linewidth]{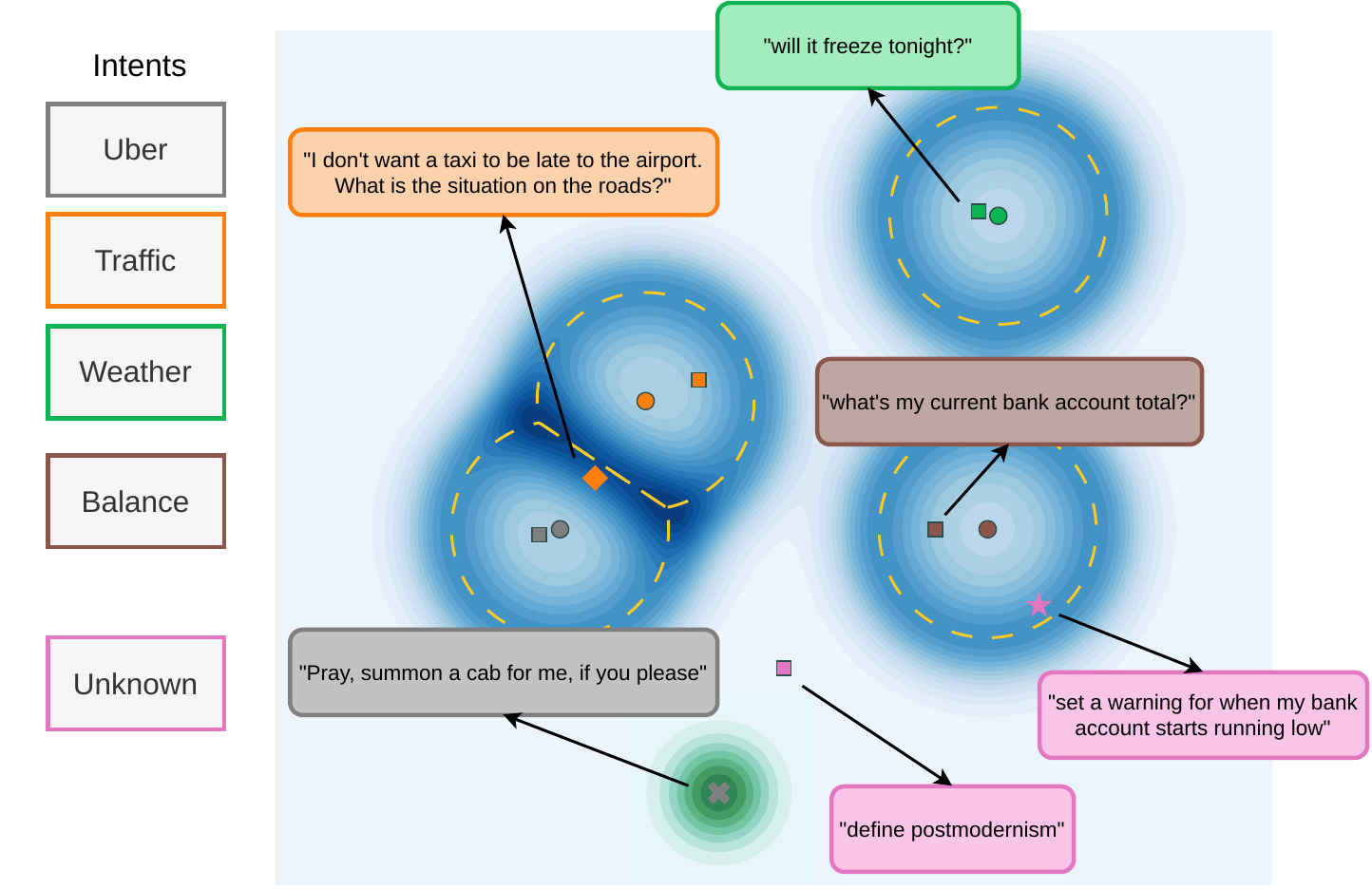}
    \centering
    \caption{An illustrative example in a two-dimensional embedding space demonstrating uncertainty estimation for intent classification.  
The gallery consists of four known intent classes (Uber, Traffic, Weather, Balance), each marked with a distinct color, while the pink color indicates embeddings of the unknown class.  
Squares represent individual text query embeddings, while circles denote class centers.  
Dashed yellow lines depict the decision boundaries among the classes.  
The intensity of the blue shading reflects uncertainty due to class ambiguity and is proportional to the entropy of $p(c|\vecZ)$, where $\vecZ\in\mathbb{R}^{2}$ is the embedding of a text query and $c$ is the intent label.  
Entropy is highest near decision boundaries between different intents.  
The $\blacklozenge$ symbol lies between the Uber and Traffic classes, exemplifying high uncertainty and highlighting the risk of misclassification for semantically overlapping queries.  
The $\bigstar$ symbol resides near the boundary of the Balance class, exhibiting high uncertainty and representing a potential false acceptance of an out-of-domain query.  
The $\boldsymbol{\times}$ symbol represents a noisy or stylistically distinct in-gallery sample whose embedding is displaced from its true class center.  
Text annotations provide concrete examples of natural language queries mapped to their respective embedding locations.  
The visualization demonstrates how spatial uncertainty helps distinguish between confident classifications, ambiguous queries, and unknown intents.}
    \label{fig:teaser}
\end{figure*}

In the image domain, discriminative embeddings are successfully used to encode raw input and drive OSR decisions~\cite{gunther2017toward}.
Moreover, constructing a Bayesian probabilistic model that incorporates feature uncertainty and the relative position of embeddings allows for a principled uncertainty estimate of system predictions~\cite{my_paper}.
The OSR task involves three types of errors: misidentification, false rejection, and false acceptance.
Figure~\ref{fig:teaser} illustrates how these error types are captured by our uncertainty scores.
Misidentification and false acceptance errors can be detected using information about the structure of the gallery.
Embeddings of such erroneous samples often lie near decision boundaries between classes, resulting in high uncertainty scores.
False rejections occur when a sample's embedding lies far from its class center.
In the image domain, this is caused by ambiguity in visual features, such as occlusion, blur, and corruption.
Probabilistic embeddings, e.g., Sphere Confidence Face (SCF)~\cite{scf}, are able to detect ambiguous images; we hypothesize that this approach can be applied to text.
Indeed, during SCF training, embeddings of ambiguous queries are mapped far from their class center, necessitating high variance assignments for such samples. 

% our contributions
In this paper, we bridge this gap by adapting a principled Bayesian uncertainty estimation framework to the text domain. 
We do not propose a new uncertainty method ab initio; rather, we adapt the Holistic Uncertainty Estimation (HolUE) framework, originally developed for biometric recognition~\cite{my_paper}, to text-based OSR systems.
We hypothesize that the sources of uncertainty identified in face recognition—gallery structure and embedding variance—are equally transferable to text embeddings derived from transformer models. 
To facilitate this research, we also construct a challenging OSR benchmark for authorship attribution.
Our extensive experiments demonstrate that capturing both sources of uncertainty significantly improves the system's ability to filter erroneous decisions compared to methods that consider only sample quality or acceptance scores.

The key contributions of our work are as follows:
\begin{itemize}
    \item We identify the primary sources of uncertainty in NLP-based OSR systems: query ambiguity and gallery structure.
    \item We adapt a principled Bayesian uncertainty score for text-based OSR systems.
    \item We release a new authorship attribution protocol based on the PAN dataset, providing a challenging OSR benchmark that reflects the dynamic nature of known author galleries.
    \item We show through extensive experiments on various text-related tasks, including intent and topic classification, that our Bayesian uncertainty score reliably detects all types of recognition errors, outperforming standard uncertainty baselines.
\end{itemize}

The remainder of this paper is organized as follows: Section 2 reviews related work in open-set text classification and uncertainty estimation.
Section 3 formalizes the background and problem statement. Section 4 details the adaptation of the uncertainty estimation method to the text domain.
Section 5 presents our experimental setup and results, and Section 6 concludes the paper.

\section{RELATED WORK}
\subsection{Open-Set Text Classification}

Open-Set Recognition (OSR) is a fundamental challenge in Open-World Machine Learning (OWML), differing from closed-set classification by the possibility of encountering previously unseen classes during inference~\cite{osti_10059464, osr_in_text_review}.
In the text domain, this problem is formalized as Open-Set Text Classification (OSTC).
Here, a system must classify a text sample into one of the known classes or reject it as unknown.
Several Natural Language Processing (NLP) tasks naturally formulate as OSTC problems, including intent classification in conversational agents~\cite{clinc150, lin-xu-2019-deep}, authorship attribution for verification~\cite{stamatatos-2017-authorship, 10.1561/1500000005, 10.1007/978-3-030-86337-1_15}, and topic recognition~\cite{yahoo_ostc} for content filtering.
For instance, in intent classification, the system must handle out-of-scope queries without forcing misclassification, while in authorship attribution, it must distinguish between known writers and impostors.
Despite the prevalence of these tasks, OSTC remains less explored than its computer vision counterparts, with many systems still operating under closed-world assumptions.

\subsection{Existing methods}
% here I list and describe methods that were used to solve OSR problem in text
Traditional approaches to OSR rely on discriminative embedding models (e.g., ArcFace~\cite{deng2019arcface}) and distance-based thresholds to separate known from unknown identities~\cite{Gnther2017TowardOF, voxblink}.
In the text domain, methods typically optimize embedding discriminability to minimize error rates such as False Acceptance or False Rejection.
Prominent techniques include Deep Open Classification (DOC), which replaces the SoftMax layer with a sigmoid layer to reduce open-space risk, and OpenMax, which utilizes meta-recognition to estimate the likelihood of unseen classes~\cite{doc, bendale2016towards}.
Other approaches involve Center-Based Similarity SVM (CBS-SVM) for incremental learning and energy-based models for out-of-distribution detection~\cite{fei-liu-2016-breaking}.
Recently, probabilistic embedding models originally developed for biometrics, such as Probabilistic Face Embeddings (PFE)~\cite{pfe} and Spherical Confidence Face (SCF)~\cite{scf}, have gained attention for their ability to capture sample quality uncertainty.
However, most existing literature focuses predominantly on improving recognition accuracy or OOD detection capabilities rather than estimating the reliability of system decisions.

\subsection{Aleatoric uncertainty}
% say about propabilistic embeddings
Uncertainty in deep learning can be decomposed into aleatoric and epistemic uncertainties~\cite{ABDAR2021243, kendall2017uncertainties, gal2016dropout}.
The aleatoric uncertainty stems from data distribution ambiguity and thus cannot be reduced.
The epistemic uncertainty is caused by the uncertainty in the parameters of the model and it can be mitigated through augmentation of the data or ensembling~\cite{lakshminarayanan2017simple}.
The main focus of our work is estimation of the aleatoric uncertainty.
Probabilistic embeddings have shown success in estimating aleatoric uncertainty in the image domain by reflecting the ambiguity inherent to a sample's features in the variance of the predicted embedding~\cite{pfe}.
For example, SCF predicts a von Mises-Fisher distribution where the concentration parameter inversely correlates with variance, serving as a quality measure.
We investigate the applicability of this approach to text tasks, conjecturing that sources of uncertainty identified in face recognition—gallery structure and embedding variance—are equally applicable to text embeddings derived from transformer models.

\subsection{Research gap and our contribution}
% TODO mention text and NLP below!

To our knowledge, no prior works have attempted to estimate the uncertainty of the Open-Set Recognition system specifically within the NLP domain.
All previous research focused predominantly on the improvement of the discriminative power of the system, e.g., enhancement of open-set classification metrics for text processing. 
In our paper, we use a fixed classification system and compare different uncertainty estimation methods in terms of their ability to detect recognition errors in natural language inputs, bridging the gap between biometric uncertainty estimation and text-based OSR.

\section{BACKGROUND}

In this section, we formalize the Open-Set Recognition (OSR) problem, discuss its specific manifestations in the text domain, and define the framework for uncertainty estimation. We distinguish between the task of improving recognition accuracy and the task of estimating the reliability of system decisions.

\subsection{Open-Set Recognition Problem Statement}

The Open-Set Recognition (OSR) problem constitutes a fundamental challenge in pattern recognition, differing from closed-set classification by the possibility of encountering previously unseen classes during inference~\cite{scheirer2013toward}.
In a typical OSR scenario, the system maintains a gallery $G$ of known classes (subjects), denoted as $G = \{g_1, \dots, g_K\}$, where each $g_i$ represents a known identity or category. 
During operation, the system receives a probe sample $x$ and must address two sequential questions: first, whether $x$ belongs to any class within the gallery $G$ (acceptance), and second, if accepted, which specific class label $i \in \{1, \dots, K\}$ should be assigned (identification).
Otherwise, the probe is rejected as unknown.

There are three distinct types of errors~\cite{face_handbook}:
\begin{enumerate}
    \item False Acceptance: An unknown sample (out-of-gallery) is incorrectly accepted as known.
    \item False Rejection: A known sample (in-gallery) is incorrectly rejected as unknown.
    \item Misidentification: A known sample is accepted but assigned an incorrect class label.
\end{enumerate}

To evaluate OSR performance, standard biometric metrics are employed. 
The False Positive Identification Rate (FPIR) measures the proportion of unknown probes incorrectly accepted, while the False Negative Identification Rate (FNIR) measures the proportion of known probes that are either rejected or misidentified~\cite{face_handbook}.
A robust OSR system aims to minimize FPIR and FNIR simultaneously.
However, in risk-sensitive applications, minimizing error rates alone is insufficient; the system must also quantify the confidence of its decisions to allow for human intervention or sample reacquisition.

\subsection{Open-Set Recognition in Text Domain}

% TODO: add a table that summarizes the identified uncertainty types for specific problem

While OSR has been extensively studied in biometric modalities such as face and voice recognition, it is equally critical in Natural Language Processing (NLP). Several text classification tasks naturally formulate as OSR problems, primarily intent classification and authorship attribution.

\textbf{Intent Classification}.
In conversational AI systems, users may issue queries that fall outside the scope of supported functionalities. 
The CLINC150 dataset~\cite{clinc150} is a standard benchmark for this task, covering 150 intent classes across 10 domains.
It includes out-of-scope queries that do not match any known intent, requiring the system to reject them rather than forcing a misclassification.
Intent galleries are defined by service capabilities, and the "unknown" class represents any request the system cannot fulfill.

\textbf{Authorship Attribution}.
This task involves verifying whether a text was written by a specific author from a known set.
We utilize the PAN dataset ~\cite{DBLP:journals/corr/abs-2112-05125, Kestemont2020OverviewOT} to construct a challenging OSR benchmark.
In contrast to intent classification, authorship attribution often involves a dynamic gallery where known authors are newly created during the testing phase.
This introduces a distinct challenge: the system must distinguish between stylistic variations of known authors and entirely new writers.

Existing literature in the text domain has predominantly focused on enhancing the discriminative power of embeddings to improve OOD detection accuracy~\cite{osr_in_text_review, 1811.08581}. 
Methods often employ distance-based thresholds or energy-based models to separate known from unknown classes. However, these approaches optimize for classification metrics (e.g., AUROC, F1) rather than providing a calibrated estimate of prediction uncertainty. Consequently, while a system may detect an outlier, it may not reliably indicate whether a specific prediction is erroneous due to ambiguity or noise.

\textbf{Topic Classification}.
Similarly, topic recognition systems often encounter articles that do not fit predefined categories, requiring robust rejection mechanisms to maintain system integrity.
In this task, the system must classify text into known subject areas or reject content belonging to unseen topics, a challenge prevalent in news aggregation and content filtering where new subjects frequently emerge.
To validate our approach across diverse text structures, we employ a benchmark consisting of three datasets: Yahoo Answers, AGNews, and DBPedia, which cover questions, news articles, and Wikipedia articles, respectively~\cite{yahoo_ostc}.
Provided OSR protocol, designates a subset of topics as known classes (forming the gallery) and treats the remaining topics as unknown (out-of-gallery).
This configuration tests the system's ability to handle semantic ambiguity between known topics while correctly identifying out-of-scope content, making it a rigorous testbed for uncertainty estimation.

We provide in the table~\ref{tab:uncertainty_types} reasons for uncertain predictions for different text tasks. 
% \begin{tabular}{>{\centering\arraybackslash}m{0.25\textwidth} 
%                 >{\centering\arraybackslash}m{0.35\textwidth} 
%                 >{\centering\arraybackslash}m{0.35\textwidth}}
\begin{table}[h]
\centering
\caption{Manifestation of Uncertainty Types in Open-Set Text Classification Tasks}
\label{tab:uncertainty_types}
\begin{tabular}{lp{0.35\textwidth}p{0.35\textwidth}}
\toprule
Task & Gallery Uncertainty & Embedding Uncertainty \\ \hline
Intent Classification & Semantic overlap between known intent classes (e.g., functionally similar requests lying near decision boundaries). & Ambiguity in user phrasing, slang, or noisy input causing high variance in the embedding position relative to the class center. \\ \hline
Authorship Attribution & Stylistic similarity between known authors and impostors, leading to ambiguous decision regions in the embedding space. & Intra-author stylistic variation due to topic shifts or context changes, displacing the embedding from the author's class center. \\ \hline
Topic Classification & Semantic proximity between known topics (e.g., Politics vs. Economy), creating dense regions with high class ambiguity. & Multi-topic documents or vague content leading to dispersed embedding distributions that do not align confidently with any single topic center. \\ 
\bottomrule
\end{tabular}
\end{table}

\subsection{Uncertainty Estimation in OSR}
Uncertainty estimation in OSR serves a different objective than OOD detection.
The goal is not merely to separate known from unknown classes, but to predict the probability of system error for any given probe~\cite{my_paper}.
A robust uncertainty estimator should assign high uncertainty scores to samples likely to result in False Acceptance, False Rejection, or Misidentification, enabling the system to filter risky decisions.

\textbf{Sources of Uncertainty.} In alignment with recent findings in biometric OSR~\cite{my_paper}, we identify two primary sources of uncertainty in text classification systems:
\begin{enumerate}
    \item \textbf{Gallery Uncertainty:} Arises from the geometric structure of the embedding space. If a probe embedding lies near the decision boundary between two known classes or near the acceptance threshold, the decision is ambiguous regardless of sample quality.
    \item \textbf{Embedding Uncertainty:} Stems from the input data quality or inherent ambiguity.
    In text, this corresponds to semantic ambiguity, noisy phrasing, or stylistic outliers that cause the embedding distribution to have high variance.
\end{enumerate}

To assess the quality of uncertainty estimation, we employ metrics that evaluate the ranking of errors rather than binary classification accuracy.
Prediction Rejection (PR) curve plots the recognition performance (e.g., F1 score) against the percentage of filtered samples~\cite{1054406, NIPS2017_4a8423d5}. 
Test samples are filtered in order of decreasing uncertainty. 
A steeper curve indicates that the uncertainty score successfully identifies erroneous samples early.
We normalize the Area Under the PR Curve (AUC) to obtain the Prediction Rejection Ratio (PRR)~\cite{fadeeva2023lmpolygraph}.
The PRR compares the performance of an uncertainty method against an Oracle (which perfectly filters errors first) and a Random baseline.
It is defined as:
$$ \text{PRR} = \frac{\text{AUC}_{\text{unc}} - \text{AUC}_{\text{random}}}{\text{AUC}_{\text{oracle}} - \text{AUC}_{\text{random}}} 
$$
where $\text{AUC}_{\text{unc}}$ is the area under the curve for an uncertainty method.
A PRR of 1 indicates perfect error detection, while 0 indicates performance equivalent to random filtering.

By utilizing PRR and filtering curves, we can directly measure the operational utility of uncertainty estimation in risk-controlled scenarios. 
This framework allows us to evaluate whether a method like HolUE can effectively combine gallery awareness and embedding variance to detect all three types of OSR errors in the text domain, addressing the gap left by prior works that focus solely on accuracy enhancement.

\section{METHODS}
In this section, we detail the proposed framework for uncertainty estimation in Open-Set Text Classification. 
We begin by defining the baseline OSR decision process, followed by the architecture used to generate probabilistic text embeddings.
We then describe the Bayesian uncertainty models (GalUE and HolUE) adapted from biometric recognition to the text domain, and finally, outline the calibration procedures used to normalize uncertainty scores.

\subsection{Baseline Solution to OSR}
The standard approach to Open-Set Recognition (OSR) relies on discriminative embeddings and distance-based thresholds~\cite{face_handbook}.
Given a probe text sample $x$, the system first encodes it into a normalized feature vector $z = f(x)$.
The system computes an acceptance score $s(x)$ based on the cosine similarity between the probe embedding and the closest gallery prototype:
$$
s(x) = \max_{c \in \{1, \dots, K\}} \mu_c^T z.
$$
A predefined threshold $\tau$ determines the acceptance decision. 
If $s(x) \geq \tau$, the probe is accepted and assigned the label $\hat{c} = \arg\max_c \mu_c^T z$.
Otherwise, the probe is rejected as unknown.
A common ad-hoc uncertainty measure for this baseline is the distance to the decision boundary~\cite{huber2022stating}:
$$
q_{\text{AccScr}}(x) = |s(x) - \tau|.
$$
Low values of $q_{\text{AccScr}}(x)$ indicate high uncertainty, as the sample lies near the acceptance threshold. 
However, this measure ignores the geometric structure of the gallery (e.g., overlapping classes) and sample quality ambiguity, which motivates our probabilistic approach.

\subsection{Probabilistic text embeddings}\label{sec:probabilistic_embeddigns}
\begin{figure*}
    \includegraphics[width=1\linewidth]{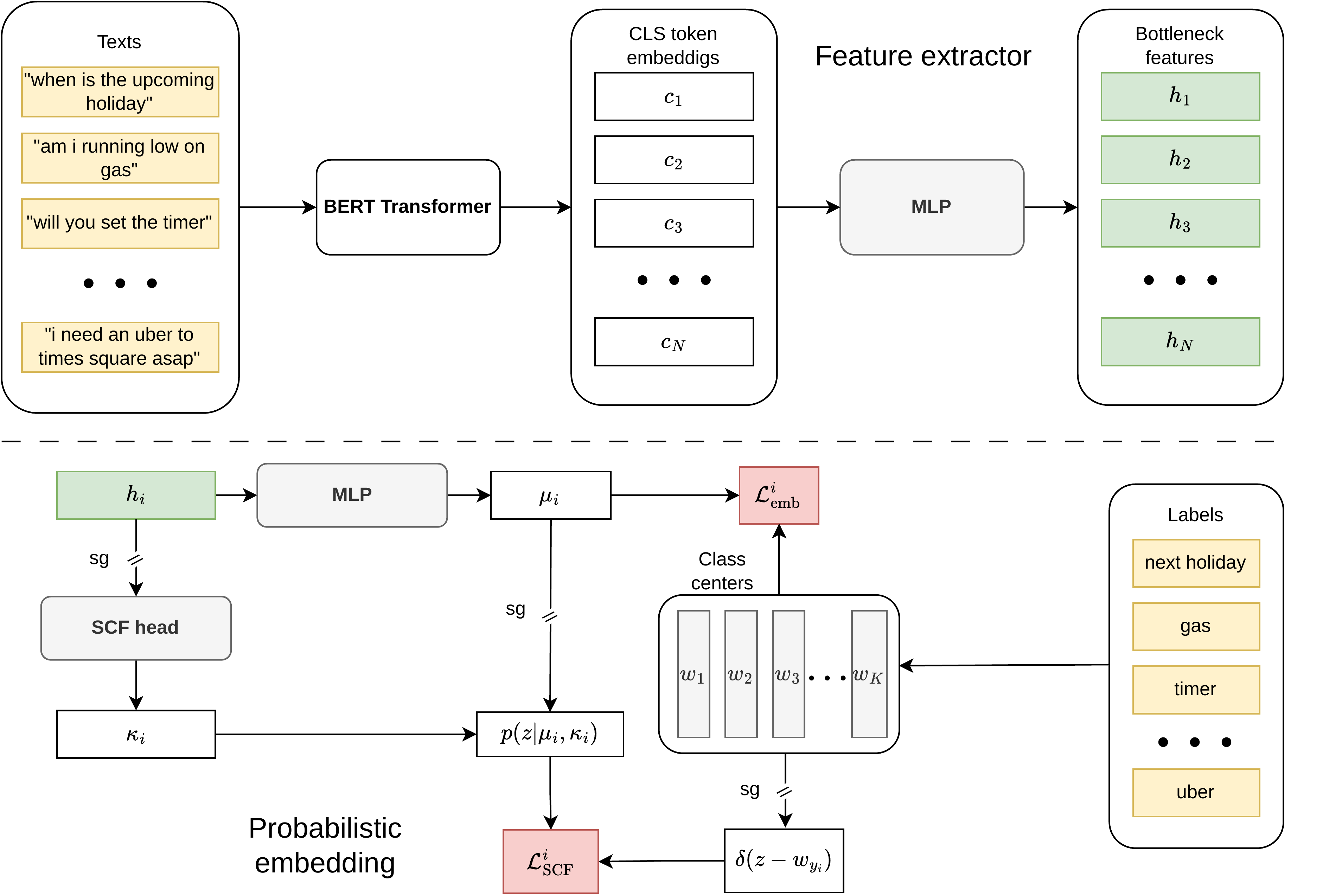}
    \centering
    \caption{Architecture of the probabilistic text embedding training. The pipeline consists of two main components:
    \textbf{Top: Feature Extraction}. Input texts are encoded using a pre-trained BERT Transformer, producing CLS token embeddings ($c_1, \ldots, c_N$).
    These embeddings are projected through an MLP into bottleneck features ($h_1, \ldots, h_N$) that serve as the input for uncertainty estimation.
    \textbf{Bottom: Probabilistic Embedding \& Loss Computation}. 
    Each bottleneck feature $h_i$ is processed through two parallel heads: (1) an MLP head that predicts the mean embedding $\mu_i$, and (2) an SCF head that predicts the concentration parameter $\kappa_i$, which characterizes embedding uncertainty.
    Together, $\mu_i$ and $\kappa_i$ define a von Mises-Fisher distribution $p(z|\mu_i, \kappa_i)$ over the hypersphere. 
    Stop-gradient (sg) operations are applied at three critical points: (i) between $h_i$ and the SCF head to prevent uncertainty estimation from affecting feature extraction, (ii) between $\mu_i$ and the probabilistic sampling to isolate the embedding loss $\mathcal{L}^i_{\text{emb}}$, and (iii) between class centers $w_k$ and the SCF loss $\mathcal{L}^i_{\text{SCF}}$ to ensure stable gallery representation.
    The framework computes two complementary losses: $\mathcal{L}^i_{\text{emb}}$ optimizes embedding discriminability by pulling $\mu_i$ toward learnable class centers, while $\mathcal{L}^i_{\text{SCF}}$ trains the concentration predictor to assign high variance to ambiguous samples (e.g., whose embedding is far from its class center).
    }
    \label{fig:backbone}
\end{figure*}

To capture uncertainty arising from text ambiguity (e.g., semantic noise, stylistic outliers), we employ probabilistic embeddings rather than deterministic point estimates. 
Our architecture, illustrated in Figure~\ref{fig:backbone}, adapts the Spherical Confidence Face (SCF) framework~\cite{scf} to transformer-based text models.
The pipeline consists of two stages:
\begin{enumerate}
    \item Feature Extraction: Input texts are encoded using a pre-trained BERT Transformer~\cite{devlin2019bert}.
    We extract the `[CLS]` token embeddings $c_{\text{CLS}}$ and project them through a Multi-Layer Perceptron (MLP) bottleneck to obtain feature vectors $h$.
    \item Probabilistic Head: The bottleneck features $h$ are processed by two parallel heads. 
    The first head predicts the mean embedding direction $\mu(\vecX) \in \mathbb{S}^{d-1}$. 
    The second head predicts a concentration parameter $\kappa(\vecX) \in \mathbb{R}^+$, which inversely correlates with variance.
\end{enumerate}

Together, $\mu(x)$ and $\kappa(x)$ define a von Mises-Fisher (vMF) distribution over the hypersphere $\mathbb{S}^{d-1}$, representing the probabilistic embedding $p(z|x)$\cite{fisher1993statistical}:
$$
p(\vecZ|\vecX) = C_d(\kappa(\vecX)) \exp(\kappa(\vecX) \mu(\vecX)^T z),
$$
where $C_d(\kappa)$ is the normalization constant. A low concentration $\kappa(\vecX)$ indicates high uncertainty regarding the sample's position in the embedding space, often caused by ambiguous or noisy text inputs. This distribution serves as the foundation for our holistic uncertainty estimation.

The system is trained in a staged manner to ensure stable uncertainty estimates.
First, bottleneck $h$ and embedding $\mu$ projections are trained together with class centers $w$ using a discriminative loss (e.g., ArcFace) to establish a structured embedding space.
In our configuration, the BERT backbone is frozen during this phase to preserve pre-trained semantic representations.
Subsequently, for uncertainty estimation training, the backbone and projection layers remain frozen.
The SCF head is then trained to predict $\mu(\vecX)$ and $\kappa(\vecX)$ using the probabilistic embedding loss.
We optimize only SCF head to prevent the uncertainty loss from altering the discriminative feature space established during the first stage.

\subsection{Bayesian OSR model}
To obtain a holistic uncertainty estimate that accounts for both embedding variance and gallery structure, we formulate the Open-Set Text Classification problem within a Bayesian probabilistic framework.
Our goal is to reconstruct the posterior class distribution $p(c|x)$ given a text sample $x$.
This distribution integrates over the embedding space $S^{d-1}$, combining the probabilistic embedding distribution $p(z|x)$ derived in Section~\ref{sec:probabilistic_embeddigns} with the gallery-aware class likelihood $p(c|z)$:
\begin{equation}\label{eq:integral}
    p(c|x) = \int_{S^{d-1}} p(c|z)p(z|x)dz
\end{equation}

where $c$ denotes the class label (intent or author), and $z \in S^{d-1}$ is the text embedding on the $d$-dimensional unit sphere. 
The differential entropy of $p(c|x)$ corresponds to the uncertainty of the model; however, to obtain a uncertainty we compute the Kullback-Leibler (KL) divergence between the posterior $p(c|x)$ and the prior class distribution $p(c)$, which is well defined for mixed probability density.

We model the gallery structure using a generative approach. By applying Bayes' rule, the probability of a class given an embedding is defined as:

$$
p(c|z) = \frac{p(z|c)p(c)}{p(z)}.
$$
We assume a mixed random variable for the class label $c \in \{1, \dots, K\} \cup (K, K+1]$. Discrete values correspond to the $K$ known classes in the gallery, while continuous values in $(K, K+1]$ represent the continuum of out-of-gallery (unknown) classes. The prior probability density function is defined as:
$$
p(c) = \frac{1 - \beta}{K} \sum_{i=1}^{K} \delta(c - i) + \beta \mathbb{I}\{c \in (K, K+1]\}
$$
where $\beta \in [0, 1]$ is the prior probability mass assigned to the unknown class continuum, $\delta$ is the Dirac delta function, and $\mathbb{I}$ is the indicator function. 
This uniform prior over the unknown space ensures that the uncertainty estimator remains sensitive to ambiguity caused by embedding shifts, preventing overconfident rejections of corrupted in-gallery samples.

For known gallery classes $c \in \{1, \dots, K\}$, we model the embedding distribution using von Mises-Fisher (vMF) distributions centered at class prototypes $\mu_c$:

$$
p(z|c) = C_d(\kappa_{gal}) \exp(\kappa_{gal} \mu_c^T z)
$$
where $\kappa_{gal}$ is a concentration hyperparameter constant for all gallery classes, and $C_d(\cdot)$ is the normalization constant. For out-of-gallery classes $c \in (K, K+1]$, embeddings are assumed to be uniformly distributed on the sphere $S^{d-1}$.

To measure uncertainty, we compute the KL-divergence between the posterior $p(c|x)$ and the prior $p(c)$.
The KL-divergence decomposes into two components, $\text{KL}_1$ (related to gallery ambiguity) and $\text{KL}_2$ (related to embedding quality/unknown probability):

\begin{align*}
    \operatorname{D}_{\mathrm{KL}}(p(c|\vecX) \| p(c)) = \underbrace{\sum_{c = 1}^K \mathbb{P}(c|\vecX) \log \frac{\mathbb{P}(c|\vecX)}{\mathbb{P}(c)}}_{\mathrm{KL}_1} +\underbrace{\int_K^{K+1} p(c|\vecX) \log \frac{p(c|\vecX)}{p(c)} dc}_{\mathrm{KL}_2}
\end{align*}
where
\begin{equation*}\label{eq:gallery_prob}
    \mathbb{P}(c|\vecX) = \int_{\mathbb{S}_{d - 1}} \frac{1 - \beta}{K}\frac{p(\vecZ|c)}{p(\vecZ)}p(\vecZ|\vecX) d\vecZ.
\end{equation*}
with reparameterization we can rewrite second term as an integral over hypersphere:
\begin{equation*}
    \mathrm{KL}_2 = \int_{\mathbb{S}_{d - 1}} \frac{1}{S_{d - 1}} \beta\frac{p\left(\vecM_c^\circ| \vecX\right)}{p\left(\vecM_c^\circ\right)} \log \frac{ p\left(\vecM_c^\circ|\vecX\right)}{ p\left(\vecM_c^\circ\right)} d \vecM_c^\circ. 
\end{equation*}

Computation of the integrals in the KL-divergence is analytically intractable.
While Monte Carlo integration is theoretically possible, we follow the approximation strategy validated in our previous work, which uses the mean embedding $\vecM_{\vecX} = \mu(\vecX)$ as a representative point for the distribution $p(\vecZ|\vecX)$:
\begin{equation*}
    \mathbb{P}(c|\vecX)\approx \frac{1-\beta}{K}\frac{p(\vecM_{\vecX}|c)}{p(\vecM_{\vecX})}.
\end{equation*}
and 
\begin{equation*}
    \mathrm{KL}_2 \approx \frac{\beta}{S_{d-1}} \frac{1}{p(\vecM_{\vecX})} \log\frac{p\left(\vecM_{\vecX}|\vecX\right)}{p\left(\vecM_{\vecX}\right)}. 
\end{equation*}
This simplification avoids stochastic noise while preserving the correlation between concentration and sample quality. 
Interestingly, $\text{KL}_2$ is proportional to the predicted concentration $\kappa(x)$, capturing the embedding variance.
To ensure numerical stability and consistency with the original HolUE framework~\cite{my_paper}, we apply temperature scaling $T$ to the posterior distribution~\cite{guo2017calibration} $p(c|\vecX)$ before computing the KL-divergence components
Subsequently, we normalize both KL components using statistics computed on a validation set and fuse them using a lightweight Multilayer Perceptron(MLP) $f_\theta$:

$$
q_{HolUE} = f_\theta\left(\frac{\mathrm{KL}_1 - \mathrm{KL}_1^{\mathrm{mean}}}{\mathrm{KL}_1^{\mathrm{std}}}, \frac{\mathrm{KL}_2 - \mathrm{KL}_2^{\mathrm{mean}}}{\mathrm{KL}_2^{\mathrm{std}}}\right)
$$
where $\mathrm{KL}^{\mathrm{mean}}$ and $\mathrm{KL}^{\mathrm{std}}$ are the mean and standard deviation of the respective KL components on the validation set. 
The MLP parameters $\theta$ are trained to optimize error detection (binary classification of error vs. correct prediction) at a fixed False Positive Identification Rate (FPIR). 
This post-processing ensures that the uncertainty score is calibrated and directly correlated with the probability of recognition error.

\section{EXPERIMENTS}
% Here I show tables and rejection plots, define metrics, and describe baseline models.
In this section, we evaluate the proposed Holistic Uncertainty Estimation (HolUE) framework within the OSTC setting.
We first describe the datasets and the specific OSR protocols constructed to simulate real-world risk-controlled scenarios. 
Subsequently, we discuss the performance of uncertainty estimation methods on these benchmarks.

\subsection{Datasets and protocols}
To validate the generalizability of HolUE beyond biometric modalities, we employ three distinct text-based tasks: authorship attribution, intent and topic classification.
These tasks represent different challenges in OSTC: authorship attribution involves distinguishing between stylistic variations of known writers versus impostors, while intent classification requires separating supported functional queries from out-of-scope (OOS) requests.
We construct strict OSR protocols for all datasets, ensuring disjoint author/intent sets between the gallery (known classes) and out-of-gallery probes (unknown classes).
\subsubsection{PAN Authorship Attribution}
For authorship attribution, we utilize the PAN-20-AV dataset ~\cite{DBLP:journals/corr/abs-2112-05125, Kestemont2020OverviewOT}. 
This dataset comprises pairs of documents labeled with author identifiers, originally designed for verification tasks. 
To adapt it for open-set identification, we reconstruct the data into an author-to-documents mapping. 
We enforce a minimum document density to ensure robust gallery construction, filtering out authors with fewer than 10 documents.

We define three disjoint splits based on author identities to simulate training, validation, and testing phases under open-set conditions:
\begin{enumerate}
    \item  Training Set consists of 4,000 authors used to train the backbone embedding model and the probabilistic heads. These authors are not visible during the OSR evaluation phase.
    \item Validation Set comprises 200 authors. We split these into two groups: 100 authors form the Gallery (known identities), and 100 authors serve as Out-of-Gallery probes (unknown identities). 
    This split is used for hyperparameter tuning and calibration of the uncertainty scores.
    \item Test Set comprises 200 authors, similarly split into 100 Gallery authors and 100 Out-of-Gallery authors.
    This set is reserved for final evaluation.
\end{enumerate}
For both Validation and Test phases, we construct the gallery by randomly selecting exactly 3 documents per known author. 
The probe set consists of all remaining documents from the Gallery authors (In-Gallery probes) and all documents from the Out-of-Gallery authors (Out-of-Gallery probes). 
This protocol challenges the system to reject stylistic variations of known authors that deviate significantly from the gallery templates while correctly identifying consistent samples.

\subsubsection{CLINC150 Intent Classification}
For intent classification, we employ the CLINC150 dataset, which covers 150 intent classes across 10 domains and includes explicit Out-of-Scope (OOS) queries that do not belong to any supported intent. 
We construct a validation and test protocols where the gallery is formed using the entire training split, such that all train samples for the particular intent class serve as a single gallery template. 
The probe set consists of the validation and test splits, which crucially include both in-scope queries belonging to the 150 known intents and OOS queries labeled as unknown. 
In this formulation, in-gallery probes correspond to in-scope queries from the known intents, while out-of-gallery probes correspond to the OOS queries.
The system must accept in-scope queries with the correct intent label while rejecting OOS queries as unknown.
\subsubsection{Topic Classification}
To further assess the robustness of our method across diverse text domains, we utilize a diverse benchmark consisting of three datasets: Yahoo Answers, AGNews, and DBPedia.
These datasets cover questions, news articles, and Wikipedia articles, respectively. 
Following the standard OSR protocol for this benchmark, we treat a subset of topics as known classes (In-Distribution) and the remaining topics as unknown (Out-of-Distribution).
For validation, we construct the Gallery from the in-distribution training set and the Probe set from the out-of-distribution training set. 
The Gallery is formed by enrolling a subset of samples from each known class, defined as the maximum of 1\% of the class count or 50 samples per class. To ensure consistent evaluation, we subsample the in-distribution training data to match the class distribution of the test split.
The Probe set contains the remaining in-distribution samples (known probes) and all out-of-distribution samples (unknown probes).
For testing, we use a fair test split where both Gallery and Probe sets are constructed from the test data.
The Gallery template size follows the same rule as validation (max(1\%, 50 samples) per known class).
Probe samples are organized into templates of size 5 to simulate identification scenarios.
This ensures that validation and test protocols maintain comparable class distributions while using disjoint data splits.

\subsection{Main results}
Table~\ref{tab:topic_classification} and Table~\ref{tab:pan_and_clinc} present the Prediction Rejection Ratios (PRR) for the topic classification, PAN authorship identification, and CLINC150 intent classification datasets.
We evaluate performance across different FPIR thresholds, filtering out 50\% of the test samples to compute the PRR.
\begin{table}
\caption{Prediction Rejection Ratios (PRR, $\uparrow$) for $F_1$ filtering curve for three topic classification datasets.
}
\label{tab:topic_classification}
\footnotesize
\centering
\setlength\tabcolsep{2pt}
\begin{tabular}{lccccccccccccccc}
\toprule
Method & \multicolumn{15}{c}{$\mathrm{FPIR}$} \\
 & $0.1$ & $0.2$ & $0.3$ & $0.4$ & $0.5$ & $0.1$ & $0.2$ & $0.3$ & $0.4$ & $0.5$ & $0.1$ & $0.2$ & $0.3$ & $0.4$ & $0.5$ \\
\midrule
 & \multicolumn{5}{c}{Yahoo Answers} & \multicolumn{5}{c}{AGNews} & \multicolumn{5}{c}{DBPedia} \\
\midrule
AccScr &  -0.18 &  0.26 & \underline{0.42} &  0.47 &  0.49 & \underline{-0.02} & \textbf{0.42} & \underline{0.55} &  0.57 &  0.58 &  0.36 & \underline{0.78} & \underline{0.84} & \underline{0.84} &  0.79 \\
SCF & \underline{0.17} & \underline{0.28} &  0.39 & \underline{0.49} & \underline{0.56} &  -0.04 &  -0.02 &  0.02 &  0.0 &  0.06 &  0.19 &  0.31 &  0.43 &  0.49 &  0.57 \\
GalUE &  -0.18 &  0.26 & \underline{0.42} &  0.47 &  0.49 & \underline{-0.02} & \textbf{0.42} & \underline{0.55} & \underline{0.58} & \underline{0.59} & \underline{0.48} & \underline{0.78} & \underline{0.84} & \underline{0.84} & \underline{0.8} \\
HolUE & \textbf{0.79} & \textbf{0.73} & \textbf{0.73} & \textbf{0.75} & \textbf{0.77} & \textbf{0.52} &  0.41 & \textbf{0.56} & \textbf{0.68} & \textbf{0.75} & \textbf{0.85} & \textbf{0.92} & \textbf{0.93} & \textbf{0.94} & \textbf{0.95} \\
\bottomrule
\end{tabular}
\end{table}

\begin{table}
\caption{Prediction Rejection Ratios (PRR, $\uparrow$) for $F_1$ filtering curve for PAN authorship attribution and CLINC150 intent classification datasets.
}
\label{tab:pan_and_clinc}
\footnotesize
\centering
\setlength\tabcolsep{2pt}
\begin{tabular}{lcccccccccc}
\toprule
Method & \multicolumn{10}{c}{$\mathrm{FPIR}$} \\
 & $0.1$ & $0.2$ & $0.3$ & $0.4$ & $0.5$ & $0.1$ & $0.2$ & $0.3$ & $0.4$ & $0.5$ \\
\midrule
 & \multicolumn{5}{c}{PAN-20-AV} & \multicolumn{5}{c}{CLINC150} \\
\midrule
AccScr &  -0.1 &  0.38 &  0.5 &  0.36 &  0.39 &  0.15 &  0.39 & \underline{0.49} & \underline{0.56} & \underline{0.59} \\
SCF & \underline{0.1} &  0.13 &  0.15 &  0.18 &  0.15 & \underline{0.52} & \underline{0.49} &  0.48 &  0.47 &  0.48 \\
GalUE &  -0.09 & \underline{0.42} & \underline{0.52} & \underline{0.44} & \underline{0.47} &  0.14 &  0.35 &  0.4 &  0.39 &  0.38 \\
HolUE & \textbf{0.36} & \textbf{0.5} & \textbf{0.55} & \textbf{0.5} & \textbf{0.51} & \textbf{0.73} & \textbf{0.64} & \textbf{0.59} & \textbf{0.58} & \textbf{0.6} \\
\bottomrule
\end{tabular}
\end{table}

The results demonstrate that HolUE consistently outperforms all baseline methods across all datasets and operating points.
On the Yahoo Answers dataset, HolUE achieves a PRR of $0.79$ at FPIR $0.1$, significantly surpassing SCF ($0.17$) and AccScr ($-0.18$). 
Similar trends are observed on AGNews and DBPedia, where HolUE maintains high PRR scores (e.g., $0.95$ on DBPedia at FPIR $0.5$), indicating robust uncertainty estimation across diverse text structures.
The negative PRR scores for AccScr on Yahoo and AGNews suggest that relying solely on the acceptance score boundary can be detrimental in topic classification tasks where class boundaries are less distinct.

On the PAN dataset, HolUE achieves a PRR of $0.51$ at FPIR $0.5$, significantly surpassing SCF (0.15) and AccScr ($0.39$). Similarly, on CLINC150, HolUE maintains a PRR above 0.58, whereas baselines fluctuate or decline at higher FPIR levels.
Notably, SCF performs poorly on the PAN dataset (PRR$\approx 0.15$), indicating that sample quality alone is insufficient to detect recognition errors in authorship attribution, where stylistic ambiguity often mimics high-quality input.
Conversely, AccScr and GalUE show stronger performance on CLINC150 but fail to match HolUE's robustness.

\begin{figure}[!t]
\centering
\includegraphics[width=0.9\linewidth]{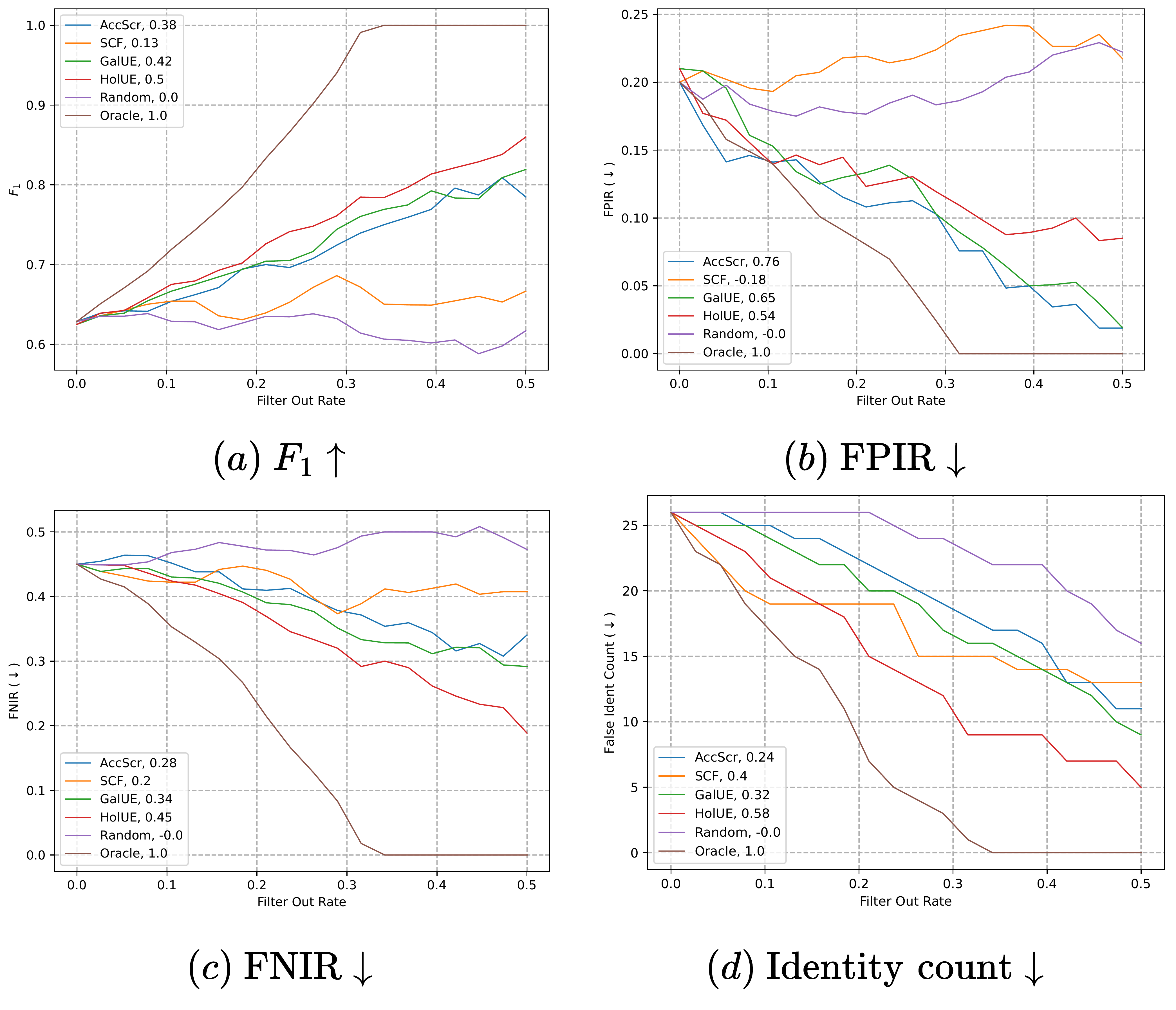}
\caption{Figure 3. Risk-Controlled Open-set Text Classification on PAN dataset starting with $\mathrm{FPIR} = 0.2$. Prediction Rejection Ratio is reported. Subplots show:(a) F1 score,(b) FNIR,(c) FPIR, and(d) False Identification Count versus filter out rate. Better to view in zoom.}
\label{fig:main_rejection_curves}
\end{figure}

To further understand the behavior of the uncertainty estimators, we analyze the risk-controlled filtering curves on the PAN dataset, as illustrated in Figure~\ref{fig:main_rejection_curves}.
The figure displays performance across four metrics: F1 score, False Negative Identification Rate (FNIR), False Positive Identification Rate (FPIR), and False Identification Count.
As shown in Figure~\ref{fig:main_rejection_curves}, HolUE demonstrates superior error filtering capability across all four metrics. 
The F1 score curve(Figure~\ref{fig:main_rejection_curves}a) shows that HolUE maintains higher recognition accuracy as samples are filtered out compared to all baselines.
Notably, the FPIR curve(Figure~\ref{fig:main_rejection_curves}c) reveals that HolUE reduces false acceptances more rapidly than AccScr and GalUE, while the FNIR curve(Figure~\ref{fig:main_rejection_curves}b) indicates better preservation of true in-gallery samples compared to SCF. 
This confirms that HolUE successfully combines the strengths of both gallery-aware(GalUE) and sample quality-aware(SCF) approaches, detecting false acceptances through gallery structure analysis and false rejections through embedding variance estimation.
The False Identification Count plot(Figure~\ref{fig:main_rejection_curves}d) further illustrates that HolUE identifies and filters erroneous decisions earlier in the rejection process.

\section{CONCLUSION}

% TODO: separate text in this section into paragraphs, add future work (link to hallucination detection and cite TOHA at least)

In this work, we addressed the critical challenge of uncertainty estimation in Open-Set Text Classification (OSTC) systems. While existing research has predominantly focused on improving recognition accuracy or out-of-distribution detection, we demonstrated that accurate uncertainty estimation is essential for building robust and trustworthy systems in risk-sensitive applications such as authorship attribution and intent classification.
We successfully adapted the Holistic Uncertainty Estimation (HolUE) framework, originally validated for biometric data, to the text domain by integrating transformer-based probabilistic embeddings with a Bayesian probabilistic model.

Our extensive experiments on the PAN authorship verification, CLINC150 intent classification, and topic classification (Yahoo Answers, AGNews, DBPedia) datasets confirm that the sources of uncertainty identified in face recognition—gallery structure and embedding variance—are equally applicable to text embeddings.
The results show that our proposed method, HolUE, consistently outperforms standard uncertainty baselines, including acceptance score-based methods (AccScr) and sample quality-based methods (SCF).
Specifically, HolUE achieved superior Prediction Rejection Ratios (PRR) across various operating points, demonstrating its ability to reliably detect all three types of Open-Set Recognition errors: false acceptance, false rejection, and misidentification.
A key insight from our study is that relying solely on sample quality or decision boundaries is insufficient for robust error detection.
High-quality text samples can still be ambiguous due to overlapping class distributions (gallery uncertainty), while noisy samples may be confidently misclassified if gallery structure is ignored. 
By combining these two sources of information through a principled Bayesian integration, HolUE provides a calibrated uncertainty score that allows systems to defer decisions to human operators when confidence is low.
Ultimately, this research bridges the gap between biometric and text-based open-set recognition, providing a domain-agnostic solution for risk-controlled deployment of machine learning systems.

In future work, we plan to extend this framework to address hallucination detection in generative language models, leveraging uncertainty estimates to identify factually inconsistent outputs.
We aim to investigate connections with recent advancements in this area, such as TOHA~\cite{toha}, to enhance reliability in open-ended text generation scenarios.

% ===== BIBLIOGRAPHY =====
% Merge both English and Russian references into single list
\bibliographystyle{unsrt} % Standard style; final formatting verified manually per journal rules
\bibliography{references}

@inproceedings{clinc150,
    title = "An Evaluation Dataset for Intent Classification and Out-of-Scope Prediction",
    author = "Larson, Stefan  and
      Mahendran, Anish  and
      Peper, Joseph J.  and
      Clarke, Christopher  and
      Lee, Andrew  and
      Hill, Parker  and
      Kummerfeld, Jonathan K.  and
      Leach, Kevin  and
      Laurenzano, Michael A.  and
      Tang, Lingjia  and
      Mars, Jason",
    editor = "Inui, Kentaro  and
      Jiang, Jing  and
      Ng, Vincent  and
      Wan, Xiaojun",
    booktitle = "Proceedings of the 2019 Conference on Empirical Methods in Natural Language Processing and the 9th International Joint Conference on Natural Language Processing (EMNLP-IJCNLP)",
    month = nov,
    year = "2019",
    address = "Hong Kong, China",
    publisher = "Association for Computational Linguistics",
    url = "https://aclanthology.org/D19-1131/",
    doi = "10.18653/v1/D19-1131",
    pages = "1311--1316",
}

@inproceedings{yahoo_ostc,
author = {Chen, Junfan and Zhang, Richong and Chen, Junchi and Hu, Chunming and Mao, Yongyi},
title = {Open-Set Semi-Supervised Text Classification with Latent Outlier Softening},
year = {2023},
isbn = {9798400701030},
publisher = {Association for Computing Machinery},
address = {New York, NY, USA},
url = {https://doi.org/10.1145/3580305.3599456},
doi = {10.1145/3580305.3599456},
booktitle = {Proceedings of the 29th ACM SIGKDD Conference on Knowledge Discovery and Data Mining},
pages = {226–236},
numpages = {11},
location = {Long Beach, CA, USA},
series = {KDD '23}
}

@inproceedings{Kestemont2020OverviewOT,
  author    = {Mike Kestemont and
               Enrique Manjavacas and
               Ilia Markov and
               Janek Bevendorff and
               Matti Wiegmann and
               Efstathios Stamatatos and
               Martin Potthast and
               Benno Stein},
  editor    = {Linda Cappellato and
               Carsten Eickhoff and
               Nicola Ferro and
               Aur{\'{e}}lie N{\'{e}}v{\'{e}}ol},
  title     = {Overview of the Cross-Domain Authorship Verification Task at {PAN}
               2020},
  booktitle = {Working Notes of {CLEF} 2020 - Conference and Labs of the Evaluation
               Forum, Thessaloniki, Greece, September 22-25, 2020},
  series    = {{CEUR} Workshop Proceedings},
  volume    = {2696},
  publisher = {CEUR-WS.org},
  year      = {2020},
  url       = {http://ceur-ws.org/Vol-2696/paper\_264.pdf},
  bibsource = {dblp computer science bibliography, https://dblp.org}
}

@article{DBLP:journals/corr/abs-2112-05125,
  author    = {Andrei Manolache and
               Florin Brad and
               Elena Burceanu and
               Antonio Barbalau and
               Radu Tudor Ionescu and
               Marius Popescu},
  title     = {Transferring BERT-like Transformers' Knowledge for Authorship Verification},
  journal   = {CoRR},
  volume    = {abs/2112.05125},
  year      = {2021},
  url       = {https://arxiv.org/abs/2112.05125},
  eprinttype = {arXiv},
  eprint    = {2112.05125},
  bibsource = {dblp computer science bibliography, https://dblp.org}
}

@article{scheirer2013toward,
  title={Toward open set recognition},
  author={Scheirer, W. J. and de Rezende Rocha, A. and Sapkota, A. and Boult, T. E.},
  journal={IEEE Transactions on Pattern Analysis and Machine Intelligence},
  volume={35},
  number={7},
  pages={1757--1772},
  year={2013},
  publisher={IEEE}
}

@book{face_handbook,
    title = {Handbook of Face Recognition},
    author = {Stan Z. Li, Anil K. Jain},
    isbn = {978-0-85729-932-1},
    year = {2011},
    publisher = {Springer London}
}

@article{osr_in_text_review,
author = {Parmar, Jitendra and Chouhan, Satyendra and Raychoudhury, Vaskar and Rathore, Santosh},
title = {Open-world Machine Learning: Applications, Challenges, and Opportunities},
year = {2023},
issue_date = {October 2023},
publisher = {Association for Computing Machinery},
address = {New York, NY, USA},
volume = {55},
number = {10},
issn = {0360-0300},
url = {https://doi.org/10.1145/3561381},
doi = {10.1145/3561381},
journal = {ACM Comput. Surv.},
month = feb,
articleno = {205},
numpages = {37}
}

@article{ABDAR2021243,
title = {A review of uncertainty quantification in deep learning: Techniques, applications and challenges},
journal = {Information Fusion},
volume = {76},
pages = {243-297},
year = {2021},
issn = {1566-2535},
author = {Moloud Abdar and Farhad Pourpanah and Sadiq Hussain and et al.},
}

@article{gunther2017toward,
  title={Toward open-set face recognition},
  author={G{\"u}nther, M. and Cruz, S. and Rudd, E. M. and Boult, T. E.},
  journal={2017 IEEE Conference on Computer Vision and Pattern Recognition Workshops (CVPRW)},
  pages={573--582},
  year={2017},
  publisher={IEEE}
}

@ARTICLE{my_paper,
  author={Erlygin, Leonid and Zaytsev, Alexey},
  journal={IEEE Access}, 
  title={Holistic Uncertainty Estimation for Open-Set Recognition}, 
  year={2026},
  volume={14},
  number={},
  pages={18868-18880},
  doi={10.1109/ACCESS.2026.3654179}}

@InProceedings{scf,
    author    = {Li, Shen and Xu, Jianqing and Xu, Xiaqing and Shen, Pengcheng and Li, Shaoxin and Hooi, Bryan},
    title     = {Spherical Confidence Learning for Face Recognition},
    booktitle = {Proceedings of the IEEE/CVF Conference on Computer Vision and Pattern Recognition (CVPR)},
    month     = {June},
    year      = {2021},
    pages     = {15629-15637}
}

@article{osti_10059464,
title = {Open Set Text Classification using Convolutional Neural Networks}, url = {https://par.nsf.gov/biblio/10059464}, journal = {International Conference on Natural Language Processing},
year = {2017},
author = {Prakhya, Sridhama and Venkataram, Vinodini and Kalita, Jugal}, }

@inproceedings{lin-xu-2019-deep,
    title = "Deep Unknown Intent Detection with Margin Loss",
    author = "Lin, Ting-En  and
      Xu, Hua",
    editor = "Korhonen, Anna  and
      Traum, David  and
      M{\`a}rquez, Llu{\'i}s",
    booktitle = "Proceedings of the 57th Annual Meeting of the Association for Computational Linguistics",
    month = jul,
    year = "2019",
    address = "Florence, Italy",
    publisher = "Association for Computational Linguistics",
    url = "https://aclanthology.org/P19-1548/",
    doi = "10.18653/v1/P19-1548",
    pages = "5491--5496",
}

@inproceedings{stamatatos-2017-authorship,
    title = "Authorship Attribution Using Text Distortion",
    author = "Stamatatos, Efstathios",
    editor = "Lapata, Mirella  and
      Blunsom, Phil  and
      Koller, Alexander",
    booktitle = "Proceedings of the 15th Conference of the {E}uropean Chapter of the Association for Computational Linguistics: Volume 1, Long Papers",
    month = apr,
    year = "2017",
    address = "Valencia, Spain",
    publisher = "Association for Computational Linguistics",
    url = "https://aclanthology.org/E17-1107/",
    pages = "1138--1149",
}

@article{10.1561/1500000005,
author = {Juola, Patrick},
title = {Authorship attribution},
year = {2006},
issue_date = {December 2006},
publisher = {Now Publishers Inc.},
address = {Hanover, MA, USA},
volume = {1},
number = {3},
issn = {1554-0669},
url = {https://doi.org/10.1561/1500000005},
doi = {10.1561/1500000005},
journal = {Found. Trends Inf. Retr.},
month = dec,
pages = {233–334},
numpages = {102}
}

@inproceedings{10.1007/978-3-030-86337-1_15,
author = {Badirli, Sarkhan and Borgo Ton, Mary and Gungor, Abdulmecit and Dundar, Murat},
title = {Open Set Authorship Attribution Toward Demystifying Victorian Periodicals},
year = {2021},
isbn = {978-3-030-86336-4},
publisher = {Springer-Verlag},
address = {Berlin, Heidelberg},
url = {https://doi.org/10.1007/978-3-030-86337-1_15},
doi = {10.1007/978-3-030-86337-1_15},
booktitle = {Document Analysis and Recognition – ICDAR 2021: 16th International Conference, Lausanne, Switzerland, September 5–10, 2021, Proceedings, Part IV},
pages = {221–235},
numpages = {15},
location = {Lausanne, Switzerland}
}

@inproceedings{deng2019arcface,
  title={Arcface: Additive angular margin loss for deep face recognition},
  author={Deng, Jiankang and Guo, Jia and Xue, Niannan and Zafeiriou, Stefanos},
  booktitle={Proceedings of the IEEE/CVF conference on computer vision and pattern recognition},
  pages={4690--4699},
  year={2019}
}

@inproceedings{voxblink,
  title={{VoxBlink2}: A {100K+} Speaker Recognition Corpus and the Open-Set Speaker-Identification Benchmark},
  author={Lin, Yuke and Cheng, Ming and Zhang, Fulin and Gao, Yingying and Zhang, Shilei and Li, Ming},
  booktitle={Proc. Interspeech 2024},
  pages={4263--4267},
  year={2024}
}

@article{Gnther2017TowardOF,
  title={Toward Open-Set Face Recognition},
  author={Manuel G{\"u}nther and Steve Cruz and Ethan M. Rudd and Terrance E. Boult},
  journal={2017 IEEE Conference on Computer Vision and Pattern Recognition Workshops (CVPRW)},
  year={2017},
  pages={573-582},
}

@InProceedings{pfe,
author = {Shi, Yichun and Jain, Anil K.},
title = {Probabilistic Face Embeddings},
booktitle = {Proceedings of the IEEE/CVF International Conference on Computer Vision (ICCV)},
month = {October},
year = {2019}
}

@inproceedings{doc,
    title = "{DOC}: Deep Open Classification of Text Documents",
    author = "Shu, Lei  and
      Xu, Hu  and
      Liu, Bing",
    editor = "Palmer, Martha  and
      Hwa, Rebecca  and
      Riedel, Sebastian",
    booktitle = "Proceedings of the 2017 Conference on Empirical Methods in Natural Language Processing",
    month = sep,
    year = "2017",
    address = "Copenhagen, Denmark",
    publisher = "Association for Computational Linguistics",
    url = "https://aclanthology.org/D17-1314/",
    doi = "10.18653/v1/D17-1314",
    pages = "2911--2916",
}

@inproceedings{fei-liu-2016-breaking,
    title = "Breaking the Closed World Assumption in Text Classification",
    author = "Fei, Geli  and
      Liu, Bing",
    editor = "Knight, Kevin  and
      Nenkova, Ani  and
      Rambow, Owen",
    booktitle = "Proceedings of the 2016 Conference of the North {A}merican Chapter of the Association for Computational Linguistics: Human Language Technologies",
    month = jun,
    year = "2016",
    address = "San Diego, California",
    publisher = "Association for Computational Linguistics",
    url = "https://aclanthology.org/N16-1061/",
    doi = "10.18653/v1/N16-1061",
    pages = "506--514"
}

@article{1811.08581,
Author = {Chuanxing Geng and Sheng-jun Huang and Songcan Chen},
Title = {Recent Advances in Open Set Recognition: A Survey},
Year = {2018},
Eprint = {arXiv:1811.08581},
Doi = {10.1109/TPAMI.2020.2981604},
}

@inproceedings{fadeeva2023lmpolygraph,
  title={LM-Polygraph: Uncertainty estimation for language models},
  author={Fadeeva, E. and Vashurin, R. and Tsvigun, A. and et al.},
  booktitle={Proceedings of the 2023 Conference on Empirical Methods in Natural Language Processing: System Demonstrations},
  year={2023}
}

@ARTICLE{1054406,
  author={Chow, C.},
  journal={IEEE Transactions on Information Theory}, 
  title={On optimum recognition error and reject tradeoff}, 
  year={1970},
  volume={16},
  number={1},
  pages={41-46},
  doi={10.1109/TIT.1970.1054406}}

@inproceedings{huber2022stating,
  title={Stating comparison score uncertainty and verification decision confidence towards transparent face recognition},
  author={Huber, M. and Terhörst, P. and Kirchbuchner, F. and Damer, N. and Kuijper, A.},
  booktitle={33rd British Machine Vision Conference 2022, BMVC 2022},
  year={2022},
  address={London, UK},
  publisher={BMVA Press}
}

@misc{toha,
Author = {Alexandra Bazarova and Aleksandr Yugay and Andrey Shulga and Alina Ermilova and Andrei Volodichev and Konstantin Polev and Julia Belikova and Rauf Parchiev and Dmitry Simakov and Maxim Savchenko and Andrey Savchenko and Serguei Barannikov and Alexey Zaytsev},
Title = {Hallucination Detection in LLMs with Topological Divergence on Attention Graphs},
Year = {2025},
Eprint = {arXiv:2504.10063},
}

@inproceedings{bendale2016towards,
  title={Towards open set deep networks},
  author={Bendale, Abhijit and Boult, Terrance E},
  booktitle={Proceedings of the IEEE conference on computer vision and pattern recognition},
  pages={1563--1572},
  year={2016}
}

@inproceedings{kendall2017uncertainties,
  title={What uncertainties do we need in bayesian deep learning for computer vision?},
  author={Kendall, Alex and Gal, Yarin},
  booktitle={Advances in neural information processing systems},
  volume={30},
  year={2017}
}

@inproceedings{gal2016dropout,
  title={Dropout as a bayesian approximation: Representing model uncertainty in deep learning},
  author={Gal, Yarin and Ghahramani, Zoubin},
  booktitle={International conference on machine learning},
  pages={1050--1059},
  year={2016}
}

@inproceedings{lakshminarayanan2017simple,
  title={Simple and scalable predictive uncertainty estimation using deep ensembles},
  author={Lakshminarayanan, Balaji and Pritzel, Alexander and Blundell, Charles},
  booktitle={Advances in neural information processing systems},
  volume={30},
  year={2017}
}

@inproceedings{guo2017calibration,
  title={On calibration of modern neural networks},
  author={Guo, Chuan and Pleiss, Geoff and Sun, Yixuan and Weinberger, Kilian Q},
  booktitle={International conference on machine learning},
  pages={1321--1330},
  year={2017}
}

@inproceedings{devlin2019bert,
  title={BERT: Pre-training of deep bidirectional transformers for language understanding},
  author={Devlin, Jacob and Chang, Ming-Wei and Lee, Kenton and Toutanova, Kristina},
  booktitle={Proceedings of the 2019 Conference of the North American Chapter of the Association for Computational Linguistics: Human Language Technologies},
  pages={4171--4186},
  year={2019}
}

@inproceedings{NIPS2017_4a8423d5,
 author = {Geifman, Yonatan and El-Yaniv, Ran},
 booktitle = {Advances in Neural Information Processing Systems},
 editor = {I. Guyon and U. Von Luxburg and S. Bengio and H. Wallach and R. Fergus and S. Vishwanathan and R. Garnett},
 pages = {},
 publisher = {Curran Associates, Inc.},
 title = {Selective Classification for Deep Neural Networks},
 url = {https://proceedings.neurips.cc/paper_files/paper/2017/file/4a8423d5e91fda00bb7e46540e2b0cf1-Paper.pdf},
 volume = {30},
 year = {2017}
}

@book{fisher1993statistical,
  title={Statistical Analysis of Spherical Data},
  author={Fisher, Nicholas I. and Lewis, Toby and Embleton, Brian J. J.},
  year={1993},
  publisher={Cambridge University Press},
  address={Cambridge, UK},
  isbn={9780521456999}
}

\end{document}